\documentclass[runningheads]{llncs}
\usepackage{graphicx}
\usepackage{xcolor}

\begin{document}
\title{Integrating Contextual Knowledge to Visual Features for Fine Art Classification 
}
\titlerunning{Contextual and Visual Features for Fine Art Classification}
%
\author{Giovanna Castellano
\orcidID{0000-0002-6489-8628} 
\and
Giovanni Sansaro
\orcidID{0000-0002-1282-6657} 
\and
Gennaro Vessio
\orcidID{0000-0002-0883-2691}}
\authorrunning{G. Castellano et al.}
%
\institute{Department of Computer Science, University of Bari Aldo Moro, Italy\\
\email{\{giovanna.castellano,gennaro.vessio\}@uniba.it}\\
\email{g.sansaro1@studenti.uniba.it}
}
\maketitle              
\begin{abstract}
Automatic art analysis has seen an ever-increasing interest from the pattern recognition and computer vision community. However, most of the current work is mainly based solely on digitized artwork images, sometimes supplemented with some metadata and textual comments. A knowledge graph that integrates a rich body of information about artworks, artists, painting schools, etc., in a unified structured framework can provide a valuable resource for more powerful information retrieval and knowledge discovery tools in the artistic domain. To this end, this paper presents \textit{$\mathcal{A}$rt$\mathcal{G}$raph}: an artistic knowledge graph based on WikiArt and DBpedia. The graph, implemented in Neo4j, already provides knowledge discovery capabilities without having to train a learning system. In addition, the embeddings extracted from the graph are used to inject ``contextual'' knowledge into a deep learning model to improve the accuracy of artwork attribute prediction tasks.

\end{abstract}
\section{Introduction}
In recent years, Knowledge Graphs (KGs) have emerged as a powerful tool for describing real-world entities and their relationships, and are increasingly used for many practical tasks, from recommendations to risk assessment \cite{hogan2021knowledge}. At the same time, the last decade has seen a remarkable range of advances in Machine Learning---and particularly in Deep Learning (DL) approaches based on neural networks \cite{lecun2015deep}---, to build ever more accurate systems in a wide range of areas, particularly computer vision and natural language processing. Combining the expressiveness of KGs with the learning ability of deep neural networks promises to develop even more effective algorithms for many \textit{downstream} tasks.

One of the many domains that can benefit from using KGs in conjunction with DL solutions is the artistic one. Leveraging DL algorithms in this domain, particularly Convolutional Neural Network (CNN) models, has already proven effective in tackling several challenging tasks, from object detection in paintings to style classification \cite{castellano2021deep}. And this success is mainly due to the growing availability of large digitized fine art collections, such as WikiArt.\footnote{https://www.wikiart.org/} However, while promising, most of the existing solutions rely solely on the visual features that a CNN can automatically extract from digital images of paintings, drawings, etc. (e.g., \cite{cetinic2018fine,sandoval2019two,strezoski2018omniart}). This inevitably leads to the neglect of an enormous amount of knowledge---already available from disparate sources---, relating to the ``context'' of each artwork. An artwork, in fact, is characterized not only by its visual appearance, but also by various other historical, social and contextual factors that place it in a much more complex and multifaceted scenario.

A promising way to harness this knowledge to improve the accuracy of art-based analytic systems is to encode the contextual information of the artworks into a KG and use an appropriate representation of the nodes in the graph, for example by means of \textit{embeddings} \cite{goyal2018graph}, as a novel, additional input to a deep learning model. Our research goes in this direction.

\paragraph{Related Work} For reasons of space, we limit our review of the related literature only to the work most directly linked to the research presented here. The interested reader can refer to our recent review article \cite{castellano2021deep} for a broader view on the computational analysis of art. 
This paper is inspired by research conducted by Garcia et al.~\cite{garcia2020contextnet}. They combined a multi-output model trained to solve attribute prediction tasks based on visual features with a second model based on non-visual information extracted from artistic metadata encoded using a KG. This model was intended to inject ``context'' information to improve the performance of the first model. The general framework was called \textit{ContextNet}. To encode the KG information into a vector representation, the popular node2vec model \cite{grover2016node2vec} was adopted. The KG was built using only the information provided by SemArt, a dataset previously proposed in \cite{garcia2018read} that provides not only artwork images and their attributes, but also artistic comments intended to achieve semantic art understanding. However, metadata are only available for artworks in the dataset, so adding a new artwork would not result in any domain information about it. In addition, the proposed graph has the artist node, which allows to connect artworks with the same artist, but without considering the relationships between artists, such as artistic influence.

\paragraph{Our Contribution}
The two limitations mentioned above can be overcome by relying on a source of knowledge external to the dataset, such as Wikipedia, which provides an enormous amount of information, even in a structured form. Furthermore, the KG could not be treated only as an adjacency matrix from which embeddings can be extracted as auxiliary information to be provided to learning models. Instead, the KG can be encoded into a NoSQL database, such as Neo4j, which can already help provide a powerful knowledge discovery framework without explicitly training a learning system. In this paper we present \textit{$\mathcal{A}$rt$\mathcal{G}$raph}, an artistic knowledge graph. The proposed KG integrates information collected by WikiArt and DBpedia, and exploits the potential of the Neo4j database management system, which provides an expressive modeling and graph query language. The proposed KG encodes a broad representation of the artistic domain, with multiple metadata and relationships between artists. Also, we propose a novel approach to inject contextual knowledge into a deep network.

\section{\textit{$\mathcal{A}$rt$\mathcal{G}$raph}}

\textit{$\mathcal{A}$rt$\mathcal{G}$raph} is a KG in the art domain capable of representing and describing concepts related to artworks. Our KG can represent a wide range of relationships, including those between artists and their works. A comparison between our proposed KG and the one presented by Garcia et al.~is provided in Table \ref{tab:comparison}. It is worth noting that, at the current stage of our research, we are only focusing on (the most popular) $300$ artists, as we are interested in a richer representation of the relationships between them and other entities.

\begin{figure}[t]
\begin{center}
\includegraphics[width=.8\linewidth]{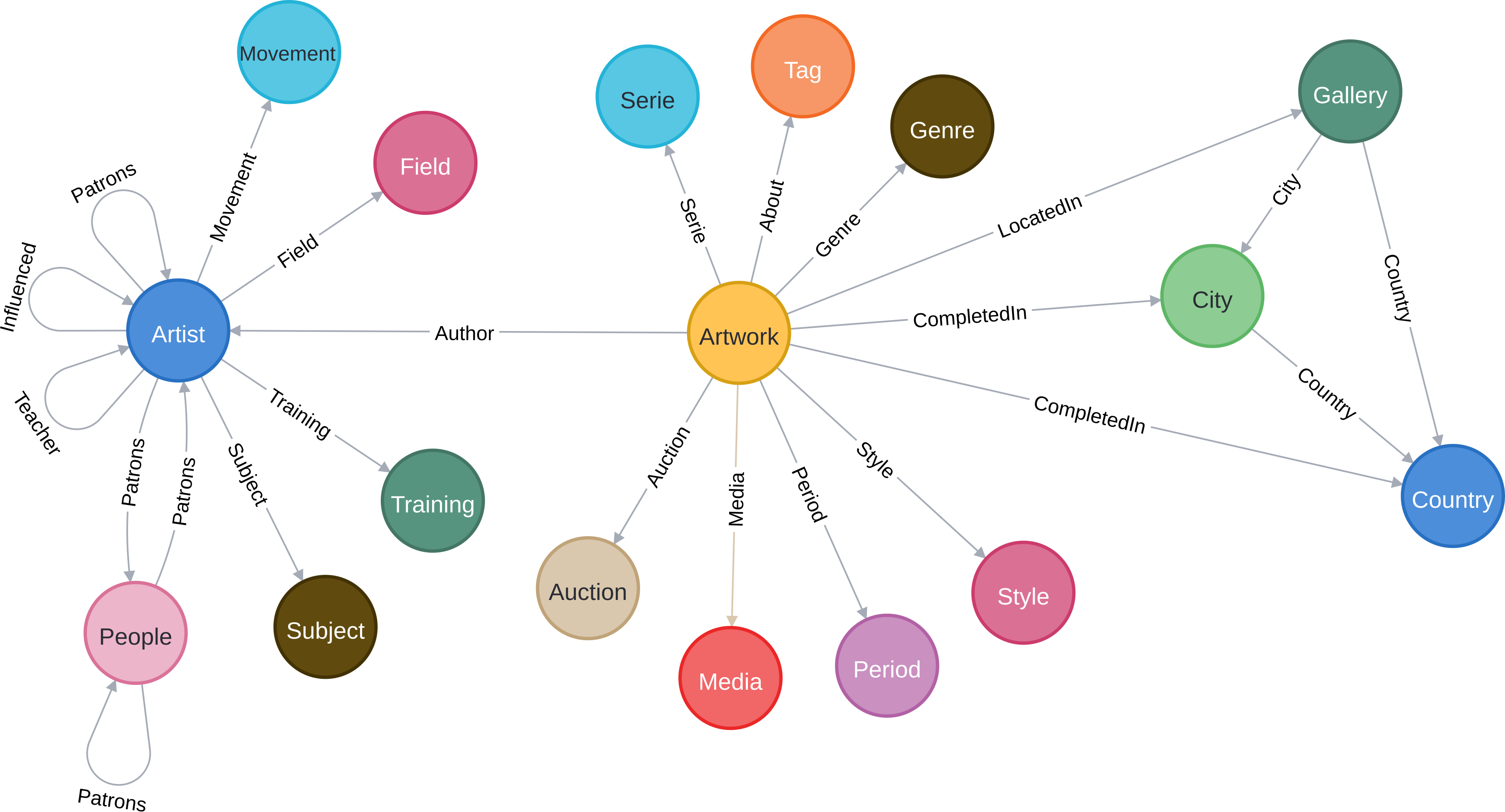}
\end{center}
\caption{Scheme of \textit{$\mathcal{A}$rt$\mathcal{G}$raph}. The nodes correspond to relevant entities in the artistic domain, while the edges represent existing relationships between them.}
\label{fig:scheme}
\end{figure}

The metadata extracted from WikiArt have been transformed into relationships and nodes mainly related to the artworks, their genre, style, location, etc. Furthermore, since WikiArt does not provide rich information about artists, each artist of our KG is linked not only to the artworks produced but also to other nodes built using RDF triples extracted from DBpedia. Extracting and integrating data from these two sources required a laborious process of data cleaning and normalization, as well as manual intervention to resolve several inconsistencies among the data. Overall, the conceptual scheme of \textit{$\mathcal{A}$rt$\mathcal{G}$raph} (represented in Fig.~\ref{fig:scheme}) includes \textit{artwork} nodes and \textit{artist} nodes: 
\begin{itemize}
    \item Each \textit{artwork} node is connected to the following nodes:  \textit{tags} (e.g., woman, sea, birds), \textit{genre} (e.g., self-portrait), \textit{style}, \textit{period}, \textit{series} (e.g., ``The Seasons'' by Giuseppe Arcimboldo), \textit{auction}, \textit{media} (e.g., paper, watercolor), the \textit{gallery} in which the artwork is located, and the \textit{city} (or \textit{country}) in which the artwork has been completed.
    \item Each \textit{artist} node is connected to the following nodes: \textit{field} (e.g., drawing, sculpture), \textit{movement} (e.g., Surrealism, Renaissance, Pop Art), \textit{training} (e.g., Accademia di Belle Arti di Firenze), \textit{Wikipedia categories} (e.g., living people, people from Florence), other artists (\textit{influences} or \textit{teaching}, and \textit{patrons}).
\end{itemize}
This structure allows the creation of a network between artists, which is useful for further analysis. In total, the resulting KG contains $74,382$ nodes and $537,883$ edges, with $300$ artists, $63,145$ artworks, $81$ genres, $49$ styles, and a huge plethora of metadata and textual comments describing them (Table \ref{tab:comparison}).

\begin{table}[t]
\begin{center}
\begin{tabular}{|l|c|c|c|c|c|c|}
\hline
KG & \# nodes & \# edges & \# artists & \# artworks & \begin{tabular}[c]{@{}l@{}}\# relations \\ btw artworks\end{tabular} & \begin{tabular}[c]{@{}l@{}}\# relations \\ btw artists\end{tabular} \\
\hline\hline
\textit{ContextNet} & $33,148$ & $125,506$ & $3166$ & $19,244$ & $7$ & $0$ \\
$\mathcal{A}rt\mathcal{G}raph$ & $74,382$ & $537,883$ & $300$ & $63,145$ & $10$ & $7$ \\
\hline
\end{tabular}
\end{center}
\caption{Comparison between our KG and the one proposed by Garcia et al.~\cite{garcia2020contextnet}. It is worth noting that, although SemArt has more than $3000$ unique artists, most of them are associated with fewer than ten artworks.}
\label{tab:comparison}
\end{table}

\begin{figure}[t]
    \centering
    \includegraphics[width=\textwidth]{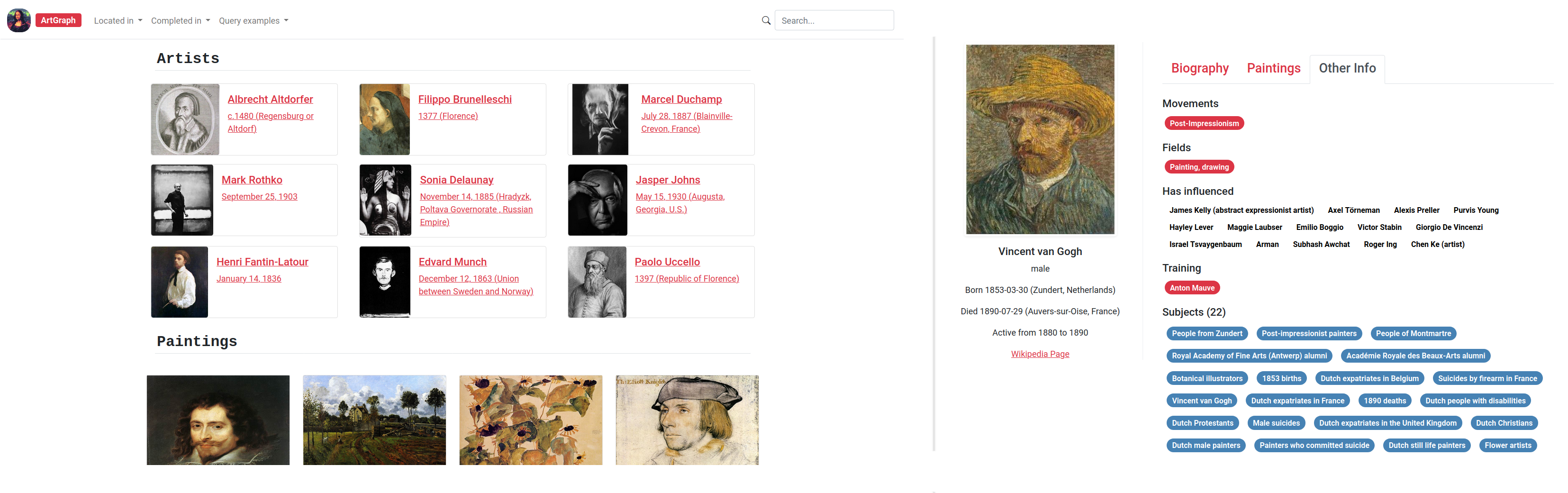}
    \caption{The home page of the  developed web interface and an example of artist page.}
    \label{fig:web}
\end{figure}

\textit{$\mathcal{A}$rt$\mathcal{G}$raph} has been implemented in Neo4j\footnote{\url{https://neo4j.com}} on an i5-10400 system, with a 2.90 GHz CPU and 16GB of RAM. We preferred Neo4j to other existing solutions as it is a native graph database that provides a powerful and flexible framework for storing and querying graph-like structures. Using Neo4j, connections between data are stored and not calculated at query time. Cypher, which is the declarative query language adopted by Neo4j, takes advantage of these stored connections to provide an expressive and optimized language for graphs to execute even complex queries extremely quickly.

To allow for a visual exploration of the graph, we have created a web interface that uses JavaScript to connect to Neo4j (Fig.~\ref{fig:web}). The goal is to provide the end user---as mentioned above, not only a generic user but especially any art historian---directly with an easy-to-use exploration tool to view the properties of an artwork or an artist. An art historian, in fact, rarely analyzes artworks as isolated creations, but typically studies how different paintings, even from different periods, relate to each other, how artists from different countries and/or periods have exercised a influence on their works, how artworks completed in one place migrated to other places, and so on. The home page randomly loads artists and artworks. Each artist is associated with a page that reports information such as the biography, the works produced, etc. We leveraged the information provided by DBpedia to show also the fields, movements, other artists who have been influenced by the current artist, and many other tags. By clicking on the buttons, the user can browse the graph interactively. The page layout of an artwork is very similar to that of an artist and reports size, period, material, etc. It is also possible to browse the artworks according to the city/country in which they were completed or are currently located. When provided by DBpedia, a textual description of the artwork is also shown.

The developed web interface can also show the results of some queries that can be particularly useful for art analysis, such as: retrieving the direct and indirect influencing connection between artists with different degrees of separation; identifying artworks that are stored in a country other than those in which they were completed; retrieving all the works that are are kept in a specific place; etc. On the tested platform, each query takes about a few tens of milliseconds. The ability to query the graph database already provides information retrieval and knowledge discovery capabilities in the art domain without having to train a learning system.

\section{Multi-Task Multi-Modal Classification}
\textit{$\mathcal{A}$rt$\mathcal{G}$raph} encodes a valuable source of contextual knowledge to integrate with visual features automatically learned by deep neural networks to develop more powerful learning models in the art domain. Several tasks, in fact, could be addressed, such as artwork attribute prediction, multi-modal retrieval and artwork captioning, which are attracting increasing interest in this domain. 

\begin{figure}[t]
\centering
\includegraphics[width=.7\textwidth]{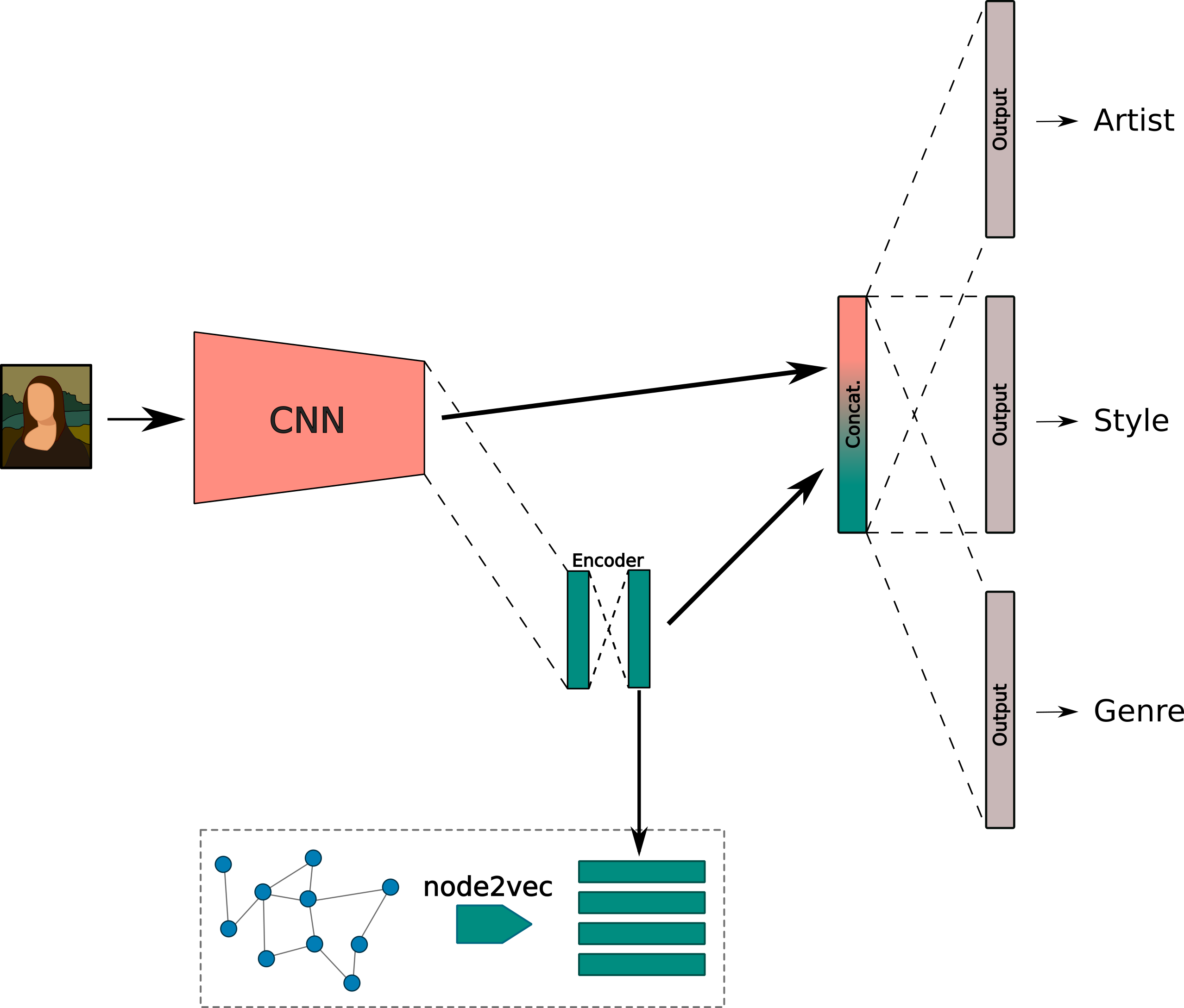}
\caption{Proposed multi-task multi-modal model. A concatenation layer receives both the contribution of visual embeddings, extracted from ResNet50 trained on digitized images of the artworks, and graph embeddings extracted from our KG, respectively. The overall network learns to minimize the error made to predict the correct artist, style and genre of a given input and, at the same time, an MSE loss to project visual and contextual features in the same multidimensional space.}
\label{fig:model}
\end{figure}

To this end, we propose a new classification model that is used in this paper to predict the artist, style and genre of a given artwork. The model is inspired by multi-modal learning: graph embeddings are extracted from \textit{$\mathcal{A}$rt$\mathcal{G}$raph} using node2vec to provide the context information of the artwork; this information is intended to improve the accuracy of visual features extracted from the artwork using a pre-trained state-of-the-art CNN, i.e.~ResNet50 \cite{he2016deep}.  
The main idea is to learn how to project the visual features extracted by ResNet50 into the context space provided by the graph embeddings. This is done by an encoder module, consisting of two fully connected layers with a tahn activation function, so that values are between $-1$ and $+1$. The training phase focuses on minimizing the mean squared error (MSE) loss between the predicted embedding $\mathbf p_j$ and the true context embedding $\mathbf u_j$, for a given artwork instance $j$:
\[
\ell_e(\mathbf p_j,\mathbf u_j) = \|\mathbf p_j - \mathbf u_j\|^2_2.
\]
Then the predicted context features are combined (by concatenation) with the visual features. Instead of adding a single output layer and learning each classification task separately, we adopt a multi-task solution. In this way, features are shared between the tasks allowing the model to simultaneously exploit the semantic correlation between them to achieve better accuracy. Given a number of task $T$ (three in our work, corresponding to the artist, style and genre classification) and a set of $N$ instances, the overall loss function is:
\[
\mathcal L = (1-\gamma) \left [ \sum_{i=1}^T \lambda_i \sum_{j=1}^N \ell_c (\mathbf z_{ij},class_{ij}) \right ] + \gamma \frac 1 N \sum_{j=1}^N \ell_e(\mathbf p_j,\mathbf u_j),
\]
where $\gamma$ weights the encoder module error, $\lambda_i$ are hyperparameters that weight the contribution of each task $i$, $\ell_e$ is the aforementioned MSE loss and $\ell_c$ is the cross-entropy loss function defined as:
\[
\ell_c(\mathbf z_j,class_j) = - \log \left ( \frac {\exp(\mathbf z_j[class_j])}{\sum_i \exp(\mathbf z_j[i])} \right ),
\]
where, for a given artwork $j$, $\mathbf z_j$ is the predicted output and $class_j$ is the true label. An overall scheme of the proposed model is shown in Fig.~\ref{fig:model}.

The experiments were conducted on Google Colaboratory. The artwork images were resized to $224 \times 224$, as required by ResNet50, and normalized using the mean and standard deviation of ImageNet. The size of the visual embeddings produced by ResNet50 (without the output layer) is 2048, while the size chosen for the node2vec embeddings is 128. As an optimizer, we used Adam with learning rate $10^{-4}$ and momentum $0.9$. The batch size was set to $32$. In addition, we empirically found the following values for: $\gamma$, which was set to $0.4$; $\lambda_{artist}$, set to $0.5$; $\lambda_{style}$, set to $0.2$; and $\lambda_{genre}$, set to $0.2$. In other words, giving more importance to the classification loss and the artist contribution to this loss generally provides better performance.

It is worth noting that graph embeddings should not be learned on the entire graph, otherwise a bias would be introduced so that the model has already seen the test entities and their connections with the rest of the graph. Instead, we assume that at test time only the visual appearance of the artwork is known to the model, but the context information learned during training has already served to allow it to generalize beyond just the visual features. For this reason, we randomly divided our graph (and consequently the image set) into three sets: 80\% for training, 10\% for validation and 10\% for test. The validation set was used to tweak the hyperparameters. Embeddings were only learned from the ``training'' graph.

The results obtained, expressed in terms of classification accuracy, are provided in Table~\ref{tab:results}. As a baseline for comparing our method, we experimented with a fine-tuned ResNet50, trained only on the digitized images. In addition, we compared our method with the \textit{ContextNet} model proposed by Garcia et al.~\cite{garcia2020contextnet}, which is also based on ResNet50 and uses graph embeddings only as a ``regularization'' signal but not as an additional input mode during training. We can see that models that incorporate contextual knowledge are better than the baseline method based only on visual features. Moreover, our model is able to better exploit context representation, with higher accuracy than \textit{ContextNet} for all three tasks.

\begin{table}[t]
\begin{center}
\begin{tabular}{|l|c|c|c|}
\hline
Method & Artist & Style & Genre \\
\hline\hline
Fine-tuned ResNet & 61.13\% & 62.65\% & 65.32\% \\
\textit{ContextNet} & 62.20\% & 62.24\% & 65.93\% \\
Proposed & \textbf{62.50\%} & \textbf{65.93\%} & \textbf{66.52\%} \\
\hline
\end{tabular}
\end{center}
\caption{Results for the artist, style and genre classification tasks.}
\label{tab:results}
\end{table}

\section{Conclusion \& Future Work}
In this paper, we have presented \textit{$\mathcal{A}$rt$\mathcal{G}$raph}, an artistic knowledge graph primarily intended to provide art historians with a rich and easy-to-use tool to perform art analysis. This effort can foster the dialogue between computer scientists and humanists that is currently sometimes lacking \cite{mercuriali2019digital}. Indeed, contrary to other works, we are not only interested in leveraging the KG information to learn classification tools, but also to help tackle knowledge discovery tasks. Humanists are interested not only in a classification model, but also in uncovering relationships, connections, trends and changes over the course of art history over time. Once stable, we will make \textit{$\mathcal{A}$rt$\mathcal{G}$raph} publicly available to provide the pattern recognition and computer vision community with a good basis for further research on automatic art analysis. 

As a future work, we want to tackle other significant tasks, such as multi-modal retrieval. Furthermore, we want to expand the proposed learning model by leveraging the Graph Convolutional Network framework, as recently done for example in \cite{vaigh2021gcnboost}.

\subsubsection*{Acknowledgements}
G.V.~acknowledges the financial support of the Italian Ministry of University and Research through the PON AIM 1852414 project.
%
%
%
\bibliographystyle{splncs04}
\bibliography{biblio}
\end{document}